\pdfoutput=1
\documentclass[11pt,a4paper]{article}
\usepackage[hyperref]{acl2021}
\usepackage{times}
\usepackage{latexsym}
\usepackage{booktabs}
\usepackage{soul}
\usepackage{array}

\usepackage{microtype}
\usepackage{graphicx}
\usepackage{multirow}

\aclfinalcopy 
\def\aclpaperid{1699} 

\newcommand{\vp}{VoxPopuli}
\newcommand{\hs}[1]{\hspace{#1\tabcolsep}}

\title{\vp: A Large-Scale Multilingual Speech Corpus for Representation Learning, Semi-Supervised Learning and Interpretation}

\author{Changhan Wang$^{\star}$, Morgane Rivi\`ere$^{\star}$, Ann Lee, Anne Wu, Chaitanya Talnikar, \\
\textbf{Daniel Haziza, Mary Williamson, Juan Pino, Emmanuel Dupoux} \vspace*{0.2cm} \\
  Facebook AI \vspace*{0.2cm} \\
  \texttt{\{changhan,mriviere,annl,annewu,talnikar,dhaziza,} \\
  \texttt{marywilliamson,juancarabina,dpx\}@fb.com}}

\date{}

\begin{document}
\maketitle

\renewcommand{\thefootnote}{$^{\star}$}
\footnotetext[1]{Equal contribution.}
\renewcommand{\thefootnote}{1}

\begin{abstract}
We introduce \vp, a large-scale multilingual corpus providing 400K hours of unlabeled speech data in 23 languages. It is the largest open data to date for unsupervised representation learning as well as semi-supervised learning. \vp~also contains 1.8K hours of transcribed speeches in 15 languages and their aligned oral interpretations into 15 target languages totaling 17.3K hours. We provide speech recognition (ASR) baselines and validate the versatility of \vp~unlabeled data in semi-supervised ASR and speech-to-text translation under challenging out-of-domain settings. The corpus is available
at \url{https://github.com/facebookresearch/voxpopuli}.
\end{abstract}

\section{Introduction}

Recent progress in speech-to-text tasks such as automatic speech recognition (ASR) and speech translation (ST) has been achieved by the development and application of unsupervised speech pre-training methods~\cite{oord2018representation,Schneider2019,baevski2020wav2vec,conneau2020unsupervised,Wu2020SelfSupervisedRI,Nguyen2020InvestigatingSP}, with semi-supervised learning (self-training)~\cite{kahn2020self,Pino2020SelfTrainingFE,zhang2020pushing,xu2020self} or a combination of both methods~\cite{xu2020self}.
This line of research leverages large amounts of unlabeled English speech data~\cite{kahn2020libri} that enable improvements in \emph{English} ASR or \emph{out-of-English} ST.
Large amounts of multilingual audio data are needed in order to achieve similar progress for multilingual ASR and ST.
Similarly, most ASR and ST research is currently conducted on the LibriSpeech~\cite{panayotov2015librispeech} and MuST-C benchmarks~\cite{cattoni2020must,di2019must}. As a result, the research community has been mostly focused on speech-to-text tasks with English as input.
While multilingual ASR~\cite{Pratap2020,ardila-etal-2020-common} and ST datasets~\cite{wang2020covost,iranzo2020europarl} have recently been made available, the amount of data available quickly drops beyond the top few high-resource languages.


Simultaneous speech translation (interpretation) has witnessed a resurgence with the applications of end-to-end encoder-decoder models. Most of the recent studies focus on text output and leverage ST corpora that are translated offline in the written form. There are differences, however, between translationese and interpretese~\citep{sridhar2013corpus,he2016interpretese}, where interpreters develop a variety of strategies to improve simultaneity. Models trained on translation corpora are unlikely to learn from these interpretation skills to achieve better quality-latency trade-offs. Finally, there has been little research~\citep{translatotron,tjandra2019speech,zhang2020uwspeech} into speech output due to the lack of open data. Existing corpora~\citep{tohyama2004ciair,bendazzoli2005approach} are either of limited size or no longer publicly available.

In this paper, we introduce \vp, a large-scale multilingual speech corpus for representation learning, semi-supervised learning and interpretation. It contains the largest open unlabeled speech data to date, totaling 400K hours in 23 languages: Bulgarian (Bg), Czech (Cs), Croatian (Hr), Danish (Da), Dutch (Nl), English (En), Estonian (Et), Finnish (Fi), French (Fr), German (De), Greek (El), Hungarian (Hu), Italian (It), Latvian (Lv), Lithuanian (Lt), Maltese (Mt), Polish (Pl), Portuguese (Pt), Romanian (Ro), Slovak (Sk), Slovene (Sl), Spanish (Es) and Swedish (Sv). \vp~also provides a total of 1.8K hours of transcribed speech in 16 languages (En, De, Fr, Es, Pl, It, Ro, Hu, Cs, Nl, Fi, Hr, Sk, Sl, Et and Lt) and their aligned oral interpretations into 15 target languages (En, De, Fr, Es, Pl, It, Ro, Hu, Cs, Nl, Fi, Sk, Sl, Lt and Da) totaling 17.3K hours.

We describe our corpus creation methodology in Section~\ref{sec:corpus_creation} and analyze the created corpus in Section~\ref{sec:data_analysis}. We provide ASR baselines and demonstrate the value of our multilingual unlabeled data as well as weakly labeled data on several non-English languages in Section~\ref{sec:experiments_and_results}.

\section{Corpus Creation}
\label{sec:corpus_creation}
\subsection{Data Acquisition}
\vp~sources data from 2009-2020 European Parliament (EP) event recordings, which include plenary sessions, committee meetings and other events. In each event, speakers give speeches in turn in different European Union (EU) languages. These speeches are partially transcribed (for plenary sessions only)
and interpreted into 24 EU languages. The interpretations are only oral without any transcription. In the following part, we refer to the original speech as ``source speech" and to the interpreted one as ``target speech".
We download audio clips for both source and target speeches from the official website\footnote{https://multimedia.europarl.europa.eu}. We also crawl the transcript, speaker information and starting/ending timestamps for each speech (for plenary sessions only) from that source, with which we later align the speech to its transcript and interpretation utterance by utterance. The acquired raw data suffers from missing audios, incomplete transcripts and inaccurate timestamps. We build data processing pipelines to segment speech paragraphs into utterances and filter out the ones with erroneous transcriptions.


\begin{table}[t]
    \centering
    \small
    \begin{tabular}{c|c|ccc|c}
    \toprule
    & Unlab. & \multicolumn{3}{c|}{Transcribed} & LM \\
    & Hrs & Hrs & Spkrs (F\%) & Tkns & Tkns \\
    \midrule
    En & 24.1K & 543 & 1313 (29.6) & 4.8M & 60.1M \\
    De & 23.2K & 282 & 531 (30.6) & 2.3M & 50.0M \\
    Fr & 22.8K & 211 & 534 (38.6) & 2.1M & 58.6M \\
    Es & 21.4K & 166 & 305 (40.6) & 1.6M & 57.4M \\
    Pl & 21.2K & 111 & 282 (23.7) & 802K & 13.6M \\
    It & 21.9K & 91 & 306 (33.8) & 757K & 52.1M \\
    Ro & 17.9K & 89 & 164 (27.6) & 739K & 10.3M \\
    Hu & 17.7K & 63 & 143 (30.3) & 431K & 13.0M \\
    Cs & 18.7K & 62 & 138 (24.9) & 461K & 13.5M \\
    Nl & 19K & 53 & 221 (39.3) & 488K & 54.6M \\
    Fi & 14.2K & 27 & 84 (56.8) & 160K & 34.5M \\
    Hr & 8.1K & 43 & 83 (33.1) & 337K & 285K \\
    Sk & 12.1K & 35 & 96 (33.8) & 270K & 13.3M \\
    Sl & 11.3K & 10 & 45 (43.9) & 76K & 12.6M \\
    Et & 10.6K & 3 & 29 (43.7) & 18K & 11.3M \\
    Lt & 14.4K & 2 & 21 (14.8) & 10K & 11.5M \\
    Pt & 17.5K & - & - & - & - \\
    Bg & 17.6K & - & - & - & - \\
    El & 17.7K & - & - & - & - \\
    Lv & 13.1K & - & - & - & - \\
    Mt & 9.1K & - & - & - & - \\
    Sv & 16.3K & - & - & - & - \\
    Da & 13.6K & - & - & - & - \\
    \midrule
    All & 384K & 1791 & 4295 & 15M & 467M \\
\bottomrule
    \end{tabular}
    \caption{Statistics for unlabeled (``Unlab.") and transcribed speech data in \vp: duration in hours (``Hrs"), number of speakers (``Spkrs"), percentage of female speakers (``F\%") and number of tokens (``Tkns"). Durations are calculated on segmented audios where leading and trailing silence is trimmed. The LM data is a combination of \vp~ transcription and sentences from EuroParl~\citep{koehn2005europarl}.}
    \label{tab:unlabeled_transcribed_stats}
\end{table}

\subsection{Data Processing}
\label{sec:data_processing}
\subsubsection{Unlabeled Speech}
We construct \vp~unlabeled set from all source and target speeches in 23 EU languages (excluding Irish because of very limited data availability). We segment full-event audios into short clips of 15-30 seconds using an energy-based voice activity detection (VAD) algorithm\footnote{https://github.com/amsehili/auditok}. Each audio clip has a maximum of 2 seconds of continuous silence, and silent clips are discarded. Around 16\% of the data is dropped after silence removal, which leads to a final overall duration of around 400K hours.

\begin{table*}[t]
    \centering
    \small
    \begin{tabular}{c|c@{\hs{1.3}}c@{\hs{1.3}}c@{\hs{1.3}}c@{\hs{1.3}}c@{\hs{1.3}}c@{\hs{1.3}}c@{\hs{1.3}}c@{\hs{1.3}}c@{\hs{1.3}}c@{\hs{1.3}}c@{\hs{1.3}}c@{\hs{1.3}}c@{\hs{1.3}}c@{\hs{1.3}}c@{\hs{1.3}}|c}
    \toprule
    \multirow{2}{*}{Source} & \multicolumn{15}{c}{Target (Oral Interpretation)} \\
    & En & De & Fr & Es & Pl & It & Ro & Hu & Cs & Nl & Fi & Sk & Sl & Lt & Da & Total \\
    \midrule
    En & - & 463 & 427 & 441 & 432 & 461 & 457 & 382 & 427 & 400 & 442 & 433 & 434 & 398 & 370 & 6.0K \\
    De & 187 & - & 196 & 204 & 214 & 217 & 198 & 205 & 214 & 196 & 217 & 208 & 218 & 164 & 179 & 2.8K \\
    Fr & 169 & 187 & - & 187 & 172 & 197 & 195 & 144 & 170 & 158 & 168 & 168 & 156 & 139 & 134 & 2.3K \\
    Es & 130 & 138 & 135 & - & 118 & 148 & 128 & 93 & 118 & 115 & 124 & 114 & 108 & 83 & 86 & 1.6K \\
    Pl & 68 & 66 & 54 & 55 & - & 67 & 55 & 43 & 67 & 42 & 55 & 62 & 57 & 50 & 34 & 775 \\
    It & 69 & 77 & 76 & 79 & 72 & - & 75 & 61 & 68 & 64 & 71 & 66 & 70 & 53 & 60 & 961 \\
    Ro & 60 & 59 & 59 & 58 & 49 & 61 & - & 38 & 50 & 43 & 48 & 50 & 46 & 38 & 29 & 688 \\
    Hu & 30 & 38 & 25 & 27 & 29 & 30 & 27 & - & 27 & 20 & 31 & 29 & 26 & 21 & 18 & 378 \\
    Cs & 39 & 35 & 29 & 30 & 36 & 32 & 31 & 23 & - & 23 & 29 & 55 & 29 & 25 & 18 & 434 \\
    Nl & 31 & 43 & 35 & 29 & 27 & 38 & 24 & 25 & 25 & - & 32 & 25 & 23 & 19 & 25 & 401 \\
    Fi & 15 & 18 & 15 & 13 & 13 & 13 & 13 & 12 & 13 & 11 & - & 14 & 12 & 11 & 9 & 182 \\
    Hr & 31 & 27 & 27 & 24 & 27 & 28 & 24 & 22 & 24 & 22 & 24 & 26 & 37 & 21 & 20 & 384 \\
    Sk & 21 & 22 & 14 & 16 & 19 & 16 & 16 & 14 & 32 & 13 & 16 & - & 17 & 13 & 10 & 239 \\
    Sl & 6 & 6 & 4 & 5 & 5 & 6 & 5 & 4 & 5 & 4 & 5 & 6 & - & 4 & 3 & 68 \\
    Lt & 1 & 1 & 1 & 1 & 1 & 1 & 1 & 1 & 1 & 1 & 1 & 1 & 1 & - & 0 & 13 \\
    \midrule
    Total & 857 & 1.2K & 1.1K & 1.2K & 1.2K & 1.3K & 1.2K & 1.1K & 1.2K & 1.1K & 1.3K & 1.3K & 1.2K & 1.0K & 995 & 17.3K \\
    \bottomrule
    \end{tabular}
    \caption{Duration statistics (hours) of aligned speech-to-speech data in \vp~between 15 source languages and 15 target languages.}
    \label{tab:s2s_align_stats}
\end{table*}

\subsubsection{Transcribed Speech}
\label{sec:transcribed_speech}
The \vp~transcribed set comes from aligning the full-event source speech audio with the transcripts for plenary sessions. Official timestamps are available for locating speeches by speaker in the full session, but they are frequently inaccurate, resulting in truncation of the speech or mixture of fragments from the preceding or the succeeding speeches. To calibrate the original timestamps, we perform speaker diarization (SD) on the full-session audio using pyannote.audio~\citep{Bredin2020} and adopt the nearest SD timestamps (by L1 distance to the original ones) instead for segmentation. Full-session audios are segmented into speech paragraphs by speaker, each of which has a transcript available.

The speech paragraphs have an average duration of 197 seconds, which leads to significant memory usage and prevents efficient parallelism (batching) during model training. We hence further segment these paragraphs into utterances with a maximum duration of 20 seconds. We leverage speech recognition (ASR) systems to force-align speech paragraphs to the given transcripts and cut the utterances by ending punctuation or the longest silence inside the sentence if it exceeds 20 seconds. The ASR systems are TDS models~\citep{TDS2019} trained with ASG criterion~\citep{collobert2016wav2letter} on audio tracks from in-house de-identified video data. The resulting utterance segments may have incorrect transcriptions due to incomplete raw transcripts or inaccurate ASR force-alignment. We use the predictions from the same ASR systems as references and filter the candidate segments by a maximum threshold of 20\% character error rate (CER).

We split the filtered utterances into train, development and test sets with disjoint speakers and target duration ratio (18:1:1). To determine the assignments, we group utterances by speaker and sort them by overall duration in ascending order. We assign the sorted groups to the test set in order until it reaches 20 speakers or the target duration (whichever comes later). The same process is repeated on the remaining utterance groups to construct the development set (with minimum 10 speakers instead). Finally, the rest of utterances make up the train set. This approach ensures higher speaker diversity in the test and development sets.

\subsubsection{Speech-To-Speech Alignment}
Even though every source speech is associated with corresponding simultaneous interpretations in target languages, considerable preprocessing and filtering is necessary to make this dataset usable. Our strategy is to align source and target at the sentence level using ASR.

We first compare the spectrogram of the source and the target speech to remove the identical parts and segment the target speech into paragraphs. These identical speech are due to either the short delay between the time the source speaker and the interpreter started, or the fact that the source language is the same as the target one, and thus no interpretation is needed. For long target paragraphs, we further segment them by silence into audio clips of at most 15 minutes long. We use the same ASR model described in Section~\ref{sec:transcribed_speech} and a language model (Section~\ref{sec:lm_data}) to decode the segmented target audio. The decoded text is also forced aligned with the target audio, so that we have the timestamps of every decoded word. 

For each source segment produced in Section~\ref{sec:transcribed_speech}, we locate all decoded words that are within a window of five seconds to its start and end. A set of candidate target segments can be generated from all possible combinations of the starting and ending decoded words. We compute the cosine similarity between the LASER representation~\citep{artetxe2019massively} of the source text and each decoded text in the candidate set to find the best target segment, i.e. the one with the highest score. We first carry out this process for all source segments, respectively, and then finetune the boundaries of overlapping target segments for consecutive source segments. Finally, a threshold of 0.75 is applied on the similarity score to filter out low-quality alignments, which can be due to ASR errors.

In addition to ASR output, we also collect human transcription on 400 hours of English target speech. The human annotators were asked to provide timestamps for each word while transcribing, and thus we can apply the same alignment process described above on human transcription and generate a set of ground truth speech-to-speech alignment data.

As a by-product from this alignment process, source text and target speech is aligned, which provides speech-to-text ``translation" data in the reversed direction. This data is weakly labeled---the label (text) may contain more information than the speech data (interpretation is likely to drop unimportant details) and hence is not exact. However, it is still useful for ST model training as an addition to labeled data.
\begin{table}[t]
    \small
    \begin{tabular}{p{0.18\linewidth} | p{0.7\linewidth}}
    \toprule
    \hspace{1000pt} Original (French) & Vous le savez tous, la for\^et recule. Toutes les deux secondes dans le monde, c'est l'\'equivalent d'un terrain de football qui est d\'etruit, c'est en un an l'\'equivalent du territoire de la Gr\`ece qui est d\'eforest\'e et c'est \'evidemment dramatique. \\
    \midrule
    \hspace{1000pt} Translation & As you all know, the forest is receding. Every two seconds, across the world, the equivalent of a football pitch is destroyed; within a year, an area the size of Greece is deforested. Clearly, this is a tragic situation. \\
    \midrule
    \hspace{1000pt} Interpretation & You all know that we are losing forests every second, the surface the size area of a football field is lost in the forest. This is really tragic. \\
    \bottomrule
    \end{tabular}
    \caption{An example from \vp~for interpretese vs. translationese. Translationese is verbatim and exact, while interpretese tends to be more general and summarizing with unimportant details dropped.}
    \label{tab:interpretataion_example}
\end{table}
\begin{table*}[t]
    \centering
    \small
    \begin{tabular}{cr|c@{\hs{1.4}}c@{\hs{1.4}}c@{\hs{1.4}}c@{\hs{1.4}}c@{\hs{1.4}}c@{\hs{1.4}}c@{\hs{1.4}}c@{\hs{1.4}}c@{\hs{1.4}}c@{\hs{1.4}}c@{\hs{1.4}}c@{\hs{1.4}}c@{\hs{1.4}}c|@{\hs{1.2}}c}
    \toprule
    & & En & De & It & Fr & Es & Pl & Ro & Hu & Nl & Cs & Sl & Fi & Hr & Sk & Avg. $\downarrow$ \\
    \midrule
    Sup. & Dev & 30.1 & 29.0 & 41.6 & 28.6 & 27.4 & 27.1 & 28.5 & 27.4 & 35.7 & 27.8 & 95.7 & 45.7 & 44.9 & 30.2 &  37.1 \\
    baseline & Test & 30.0 & 29.3 & 45.2 & 30.5 & 31.4 & 25.6 & 27.7 & 27.9 & 38.3 & 27.7 & 96.5 & 41.6 & 40.2 & 32.7 & 37.5 \\
    \midrule
    VP-10K & Dev & 15.5 & 17.2 & 19.1 & 13.9 & 8.6 & 12.8 & 8.3 & 11.5 & 18.5 & 11.1 & 20.6 & 21.1 & 15.6 & 10.4 & 14.6 \\
    + FT & Test & 16.2 & 16.2 & 21.5 & 15.4 & 11.0 & 12.5 & 9.4 & 12.0 & 19.7 & 11.8 & 26.1 & 17.1 & 14.1 & 11.1 & 15.3 \\
    \bottomrule
    \end{tabular}
    \caption{\textbf{\vp~ASR baselines and in-domain unsupervised pre-training.} We report \vp~dev and test WER for languages with $\ge$10 hours of data. Top: supervised monolingual Transformer baselines. Bottom: wav2vec 2.0 \emph{Base} model pre-trained on 10K-hour \vp~unlabeled data (23 languages) and fine-tuned on \vp~ASR data. As we can see, pre-training with in-domain unlabeled data substantially improves performance especially for low-resource languages.}
    \label{tab:vp_asr_eval}
\end{table*}
\subsubsection{Language Modeling Data}
\label{sec:lm_data}
To train language models (LM) for ASR decoding, we combine \vp~transcription in the training set with the EuroParl corpus~\citep{koehn2005europarl}, which is from the proceedings of the European Parliament from 1996 to 2011. To process the EuroParl data, we first apply the sentence segmentation tool provided with the corpus. We remove all texts in the parentheses, replace hyphens and slashes with space, and remove all other punctuation except apostrophes. All digits are converted into words, and all texts are normalized into lowercase. Table~\ref{tab:unlabeled_transcribed_stats} shows the statistics of the LM data.

\section{Data Analysis}
\label{sec:data_analysis}

\paragraph{Unlabeled speech} As we can see from Table~\ref{tab:unlabeled_transcribed_stats}, \vp~has a total of 400K hours of unlabeled data well-distributed across 23 EU languages, resulting in 8K-24K hours of data for each language. This ensures adequate data on languages with lower ASR resource, which are likely to benefit more from semi-supervised learning. It also facilitates multilingual model training since there is not much data imbalance and little need for tuning data sampling strategy.

\paragraph{Transcribed speech} The \vp~transcribed data contains 16 languages totaling 1.8K hours and 4.3K speakers, whose detailed statistics can be found in Table~\ref{tab:unlabeled_transcribed_stats}, including duration (hours) by language, number of speakers, percentage of female speakers and number of tokens. The data distribution is imbalanced and reflects the natural distribution of the number of native speakers. The remaining 7 languages (Pt, Bg, El, Lv, Mt, Sv and Da) are not covered due to either limited data volume or the availability of processing pipelines.

\paragraph{Speech-to-speech alignment} The statistics of the speech-to-speech alignment between all source languages and 15 target languages are shown in Table~\ref{tab:s2s_align_stats}. Compared with the total amount of data available for each source language (``Transcribed hours" in Table~\ref{tab:unlabeled_transcribed_stats}), we obtain target alignments for more than $70\%$ of the source sentences in En, De, Fr, Es and It, more than $50\%$ for Pl, Ro, Cs, Nl and Hr, and the rest has at least $40\%$ of source segments aligned. To examine the quality of our ASR system, we align the ASR output with the human transcription we collect on English target speech and see a word error rate (WER) of 31.7. With the human transcription, we can produce ground truth speech-to-speech alignment data that is 1.1 times larger than the size of the alignment data created from using ASR output, indicating that around $12\%$ of the low-quality alignments are filtered due to ASR errors. If we compare the ASR-based and the ground truth alignment data, there is on average a 0.75-second shift in the target segment boundaries.

\paragraph{Interpretese vs. translationese} We exemplify the differences between simultaneous oral interpretation and offline written translation using \vp~in Table~\ref{tab:interpretataion_example}. The latter is verbatim and exact compared to the original speech, while the former tends to be more general and summarizing with unimportant details dropped. Human interpreters regularly apply these tactics to make better quality-latency trade-offs. Speech-to-speech translation models may benefit from these tactics if they are trained on interpretation data that \vp~provides.

\section{Experiments \& Results}
\label{sec:experiments_and_results}
We provide \vp~ASR baselines and validate the versatility of \vp~unlabeled data in unsupervised representation learning and semi-supervised learning for ASR as well as ST. We also evaluate the quality of speech-to-speech alignment indirectly via the weakly labeled ST data it produces.

\begin{table}[t]
    \centering
    \small
    \tabcolsep=0.16cm
    \begin{tabular}{r|c|c|c|c}
    \toprule
    & \multicolumn{4}{c}{Within/Across Speaker $\downarrow$} \\
    & En & Fr & Zh & Std. $\downarrow$ \\
    \midrule
    MFCC & 12.1/23.4 & 12.6/25.5 & 11.5/21.3 & - \\
    Sup.$^\dagger$ & 6.2/8.0 & 8.7/10.8 & 7.9/10.3 & - \\
    LL-6K$^\ddagger$ & 4.5/6.2 & 8.4/12.7 & 8.2/8.2 & 1.8/2.7 \\
    \midrule
    \multicolumn{5}{l}{\textit{\vp}} \\
    \midrule
    En-500 & 6.9/9.9 & 9.6/14.5 & 8.7/9.7 & 1.1/2.2 \\
    Fr-500 & 8.1/12.1 & 9.1/13.8 & 9.2/10.1 &  0.5/1.5 \\
    En+Fr-500 & 6.9/9.8 & 9.0/13.1 & 8.6/9.6 & 0.9/1.6 \\
    \bottomrule
    \end{tabular}
    \caption{\textbf{Phoneme discriminability of unsupervised features across languages.} We report ABX discriminability score on the 10s test set from ZeroSpeech 2017$^\dagger$ for English (``En"), French (``Fr") and Mandarin (``Zh"). We compare our models with the MFCC baseline, the supervised topline and the state-of-the-art monolingual (English) model$^\ddagger$. We measure the generality of the representations by standard deviation (``Std.") of the scores across the 3 languages. We see that multilingual representations generalize better and are more robust on unseen languages. $^\dagger$~\citet{dunbar2017zero}. $^\ddagger$~\citet{riviere2020unsupervised_wild}.}
    \label{tab:zerospeech17}
\end{table}

\begin{table*}[t]
    \centering
    \small
    \tabcolsep=0.16cm
    \begin{tabular}{r@{\hs{1.2}}|c@{\hs{1.2}}c@{\hs{1.2}}c@{\hs{1.2}}c@{\hs{1.2}}|c@{\hs{1.4}}c@{\hs{1.4}}c@{\hs{1.4}}c@{\hs{1.4}}c@{\hs{1.4}}|c@{\hs{1.4}}c@{\hs{1.4}}c@{\hs{1.4}}c@{\hs{1.4}}c@{\hs{1.4}}|c@{\hs{1.2}}c}
    \toprule
    & PT & PT & \multicolumn{2}{c|}{Langs.} & \multicolumn{5}{c|}{PER $\downarrow$ (\vp~Langs.)} & \multicolumn{5}{c|}{PER $\downarrow$ (Other Langs.)} & \multicolumn{2}{c}{PER} \\
    & Domain & Hours & In & Out & Es & Fr & It & Nl & Sv & Ky & Ru & Tr & Tt & Zh & Avg. $\downarrow$ & Std. $\downarrow$ \\
    \midrule
    m-CPC$^\dagger$ & Out & 60K & 0 & 1 & 36.4 & 44.3 & 37.8 & 43.1 & 46.5 & 37.5 & 42.4 & 45.7 & 40.6 & 53.2 & 42.7 & 4.8 \\
    \midrule
    \multicolumn{10}{l}{\textit{wav2vec 2.0 Base (95M)}} \\
    \midrule
    XLSR-Mono$^\ddagger$ & In & $<$0.4K & 1 & 0 & \textbf{6.8} & 10.4 & 10.9 & 37.4 & 63.6 & 29.6 & 11.6 & 44.0 & 21.4 & 31.4 & 26.7 & 17.2 \\
    XLSR-10$^\ddagger$ & In & 1.4K & 10 & 1 & 9.4 & 13.4 & 13.8 & 16.3 & 21.0 & 8.6 & 11.2 & \textbf{11.7} & 8.3 & 24.5 & 13.8 & 5.1 \\
    VP-Mono-5K & Out & 4.5K & 1 & 0 & \textbf{6.8} & \textbf{8.6} & \textbf{7.5} & \textbf{9.7} & \textbf{9.3} & - & - & - & - & - & - & - \\
    VP-10K & Out & 10K & 5 & 18 & 8.5 & 11.9 & 11.0 & 13.6 & 15.0 & 10.9 & 12.4 & 13.1 & 8.8 & 19.3 & 12.5 & 3.0 \\
    VP-100K & Out & 100K & 5 & 18 & 7.6 & 10.3 & 9.7 & 12.2 & 13.0 & 9.4 & \textbf{10.7} & \textbf{11.7} & \textbf{8.0} & \textbf{17.5} & \textbf{11.0} & \textbf{2.7} \\
    \midrule
    \multicolumn{10}{l}{\textit{wav2vec 2.0 Large (317M)}} \\
    \midrule
    XLSR-10$^\ddagger$ & In & 1.4K & 10 & 1 & 7.9 & 12.6 & 11.7 & 14.0 & 20.6 & 7.0 & 9.3 & 9.7 & 7.2 & 22.8 & 12.3 & 5.2 \\
    XLSR-53$^\ddagger$ & In+Out & 56K & 10 & 43 & \textbf{2.9} & \textbf{5.0} & 6.7 & \textbf{5.8} & 12.2 & \textbf{6.1} & \textbf{8.1} & \textbf{7.1} & \textbf{5.1} & 18.3 & \textbf{7.6} & 4.2 \\
    VP-Mono-5K & Out & 4.5K & 1 & 0 & 5.5 & 7.0 & \textbf{6.1} & 7.2 & \textbf{6.3} & - & - & - & - & - & - & - \\
    VP-10K & Out & 10K & 5 & 18 & 6.3 & 8.9 & 7.9 & 9.3 & 9.7 & 9.3 & 9.2 & 11.3 & 7.6 & 18.8 & 9.8 & 3.2 \\
    VP-100K & Out & 100K & 5 & 18 & 5.4 & 7.7 & 6.5 & 8.0 & 8.3 & 8.5 & 8.0 & 9.8 & 6.9 & \textbf{17.3} & 8.6 & \textbf{3.1} \\
    \bottomrule
    \end{tabular}
    \caption{\textbf{Few-shot ASR with out-of-domain out-of-language unsupervised pre-training.} We adopt the Common Voice (CV) few-shot phoneme recognition setup$^\dagger$ and report test PER (phone error rate).
    Our wav2vec 2.0 models are pre-trained on \vp~(out-of-CV-domain) either with 4.5K-hour monolingual data (``VP-Mono-5K") or 10K-hour/100K-hour multilingual data (``VP-10K" and ``VP-100K"). Pre-training languages may include the ones being evaluated (``In") and others (``Out"). Our models outperform XLSR-Mono and XLSR-10 (same architecture as ours but using in-domain CV data) on most languages with out-of-domain and (partially) out-of-language pre-training. Our best model (VP-100K Large) performs competitively to XLSR-53, which leverages 52K-hour out-of-CV-domain data in addition to the CV data. $^\dagger$~\citet{riviere2020unsupervised} $^\ddagger$~\citet{conneau2020unsupervised}}
    \label{tab:cv_eval_per}
\end{table*}
\begin{table}[t]
\centering
\small
\begin{tabular}{r|c@{\hs{1.2}}c@{\hs{1.2}}c@{\hs{1.2}}|c@{\hs{1.2}}c@{\hs{1.2}}c}
\toprule
 & \multicolumn{3}{c|}{Train Hours} & \multicolumn{3}{c}{Test WER $\downarrow$} \\
  & De & Fr & Es & De & Fr & Es \\
\midrule
Baseline$^{\dagger}$ & 1582 & 787 & 660 & 12.8 & 19.4 & 16.5 \\
\midrule
VP-50K & 314 & 364 & 203 & 17.0 & 18.8 & 11.9 \\
+ LM & (20\%) & (46\%) & (31\%) & \textbf{7.8} & \textbf{9.6} & \textbf{10.0} \\
\bottomrule
\end{tabular}
\caption{\textbf{ASR with out-of-domain unsupervised pre-training and less supervision.} We report test WER on Common Voice (CV). Top: supervised baseline trained on the combination of an extended CV train set and several other corpora (decoding with LM). Bottom: our wav2vec 2.0 \emph{Base} model pre-trained on 50K-hour \vp~data (out-of-CV-domain) and fine-tuned on the standard CV train set (a subset of the baseline's one). We optionally use 4-gram LMs trained on CV for decoding. Our model outperforms the baseline (even without LM) while using less supervised train data. $^{\dagger}$Deepspeech Polyglot.}

\label{tab:wer_scores_main_cv}
\end{table}

\begin{table*}[t]
    \centering
    \small
    \begin{tabular}{r|cc|cc|cc||cc|cc|cc}
    \toprule
     & \multicolumn{2}{c|}{Fr$\rightarrow$En $\uparrow$} & \multicolumn{2}{c|}{Es$\rightarrow$En $\uparrow$} & \multicolumn{2}{c||}{De$\rightarrow$En $\uparrow$} & \multicolumn{2}{c|}{Fr $\downarrow$} & \multicolumn{2}{c|}{Es $\downarrow$} & \multicolumn{2}{c}{De $\downarrow$}  \\
     \midrule
     Train hours (EP+CV) & \multicolumn{2}{c|}{38+264} & \multicolumn{2}{c|}{32+113} & \multicolumn{2}{c||}{42+184} & \multicolumn{2}{c|}{38+264} & \multicolumn{2}{c|}{32+113} & \multicolumn{2}{c}{42+184} \\
     Test set & EP & CV & EP & CV & EP & CV & EP & CV & EP & CV & EP & CV \\
    \midrule
    (Cascaded) Baseline$^{\dagger}$ & 25.4 & 27.6 & 26.5 & 27.4 & 21.3 & 21.0 & 24.3 & 18.3 & 15.0 & 21.4 & 19.8 & 16.0 \\
    Our end-to-end baseline & 24.5 & 27.0 & 20.5 & 26.6 & 17.5 & 20.0 & 20.8 & 18.8 & 17.2 & 14.1 & 23.2 & 18.4 \\
    With 800h self-training & 26.7 & 28.6 & 22.4 & 26.8 & 18.8 & \textbf{20.1} & 19.5 & 17.3 & 15.6 & 13.7 & 21.8 & 17.5 \\
    With 3000h self-training & \textbf{27.4} & \textbf{28.9} & \textbf{22.7} & \textbf{27.3} & \textbf{19.6} & 20.0 & \textbf{19.0} & \textbf{17.0} & \textbf{15.3} & \textbf{13.2} & \textbf{21.4} & \textbf{17.3} \\
    \midrule
    400h weakly labeled & 22.9 & 10.1 & 22.2 & 10.9 & 18.0 & 8.8 & \\
    + labeled & \textbf{31.1} & \textbf{30.3} & \textbf{28.4} & \textbf{29.7} & \textbf{24.4} & \textbf{23.4} & \\
    \bottomrule
    \end{tabular}
    \caption{\textbf{ST and ASR using \vp~data for self-training or weak supervision.} Left: test BLEU for ST models. Right: test WER for ASR models. We evaluate in-\vp-domain performance with EuroParl-ST (EP) and the out-of-domain performance with CoVoST 2 (CV). We combine both corpora to train our baseline and pseudo-label 3K-hour monolingual \vp~unlabeled data for self-training.
    For ST training with weak supervision, we combine EP, CV and 300h weakly labeled data from \vp. Both approaches for leveraging \vp~data improve in-domain (EP) and out-of-domain (CV) performance simultaneously. $^\dagger$~EP baselines from \citet{iranzo2020europarl} and CV baselines from \citet{wang2020covost}.}
    \label{tab:st_self_training_eval}
\end{table*}

\subsection{Experimental Setup}
For representation learning, we perform speaker diarization before VAD-based segmentation so that each utterance contains exactly one speaker. We augment the data with time dropout, pitch modification and reverberation~\citep{kharitonov2020data} during model training.

For non-wav2vec models, we extract 80-dimensional log-mel filterbank speech features with 25ms windows size and 10ms shift. We apply per-utterance CMVN (cepstral mean and variance normalization) to the extracted features. For GPU memory efficiency, we remove training samples that have more than 60 seconds of speech or have more than 1024 characters.

We train wav2vec 2.0~\citep{baevski2020wav2vec} models with original hyper-parameter settings using fairseq~\citep{ott2019fairseq}, except for Table~\ref{tab:wer_scores_main_cv} where we use wav2letter~\citep{pratap2018w2l} and follow~\citet{talnikar2020joint} to do finetuning using both supervised CTC~\citep{graves2006ctc} loss and unsupervised wav2vec 2.0 loss. The largest model (``VP-100K") takes 10 days on 128 V100 GPUs for 1M updates. For non-wav2vec models, we train Transformer~\citep{NIPS2017_3f5ee243} with cross-entropy criterion using fairseq S2T~\citep{wang2020fairseqs2t}. For Section~\ref{sec:asr_baselines} and Section~\ref{sec:unsupervised_pretraining}, we use phoneme vocabularies for models that we evaluate with PER (phone error rate) and character vocabularies for the other. For Section~\ref{sec:self_training}, we use Unigram~\cite{kudo-richardson-2018-sentencepiece} vocabularies with 2K subwords for all models. To improve ST model training, we pre-train the encoder on the LibriSpeech~\citep{panayotov2015librispeech} ASR task.

We use the best checkpoint by validation loss for evaluation, except for Section~\ref{sec:self_training} where we average the 10 best checkpoints.
We build n-gram language models for decoding (when specified) using KenLM~\citep{heafield2011kenlm}.

\subsection{Speech Recognition (ASR) Baselines}
\label{sec:asr_baselines}

We provide monolingual Transformer baselines for the 14 languages that have more than 10 hours of transcribed data (see Table~\ref{tab:unlabeled_transcribed_stats}). Both development and test WER are reported in Table~\ref{tab:vp_asr_eval}. We see that several low-resource languages (Fi, It, Hr, Sk and Sl) suffer from high recognition errors ($>$40\% WER) due to the lack of training data. Even the highest resource one (En) has a high WER of around 30\%.

\subsection{Unsupervised Representation Learning}
\label{sec:representation_learning}
We follow the setting in~\citet{riviere2020unsupervised} to evaluate unsupervised speech representations by phoneme discriminability on 3 languages (English, French and Mandarin), and report ABX discriminability score~\citep{schatz2013evaluating} on the 10s test set from ZeroSpeech 2017~\citep{dunbar2017zero}. Standard deviation (``Std.") of the scores across the 3 languages is also reported as a measure for the generality of the representations. 
As previous studies focus on monolingual representations, we explore multilingual representations and examine their generality across languages.
We train CPC-based models~\citep{riviere2020unsupervised_wild} on 500-hour English and 500-hour French unlabeled data from \vp, respectively. And we combine English and French data with 50\% sampling (so that the total duration remains the same) for the multilingual setting.
We observe from Table~\ref{tab:zerospeech17} that the multilingual model (``En+Fr-500") performs comparably to the monolingual ones (``En-500" and ``Fr-500") on their seen languages and performs better on unseen language (``Zh"). Its scores vary less across languages (lower ``Std.") compared to ``En-500". The variance of the scores is comparable to ``Fr-500" while the average is lower. We conclude that multilingual representations generalize better across languages and are more robust on unseen languages. For quick exploration, we leverage only part of the \vp~unlabeled data and leave the validation on more data to future work.


\subsection{Semi-Supervised Learning}
We explore two semi-supervised learning settings for the application of \vp~unlabeled data: unsupervised pre-training followed by supervised fine-tuning for ASR and self-training for ASR as well as ST.
\subsubsection{ASR with Unsupervised Pre-Training}
\label{sec:unsupervised_pretraining}
Self-supervised (unsupervised) pre-training such as wav2vec 2.0~\citep{baevski2020wav2vec} substantially reduces the need of labeled data in ASR. Furthermore, multilingual pre-training~\citep{conneau2020unsupervised} allows cross-lingual transfer, which brings extra gains especially to low-resource languages. Pre-training wav2vec 2.0 models is, however, resource-intensive and hence re-training models for each task with different domains is impractical. With the large-scale multilingual data in \vp,
we explore if scaling multilingual pre-training can take us towards the one-model-fits-all paradigm by alleviating the impacts of domain or language mismatch between pre-training and fine-tuning. We train wav2vec 2.0 models~\footnote{wav2vec 2.0 \emph{Base} (95M) unless specified otherwise.} on 10K-hour, 50K-hour and 100K-hour \vp~data in 23 languages (denoted as ``VP-10K", ``VP-50K" and ``VP-100K", respectively). We also train models with 4.5K-hour monolingual data (denoted as ``VP-Mono-5K") for comparison. For quick verification, we use only part of the \vp~unlabeled data for pre-training. We leave training the models on the full 400K-hour data to future work, which is supposed to achieve even better performance.

\paragraph{In-domain pre-training} We examine the conventional in-domain pre-training setting on the \vp~ASR benchmark. We evaluate the VP-10K model, where the pre-training data is filtered so that it has no overlaps with the transcribed development and test set. From table \ref{tab:vp_asr_eval}, we see that pre-training using unlabeled data brings significant gains to all the languages (average 59\% test WER reduction). The gains are most significant on the low-resource languages, where improvements are qualitative (for example, from nearly 100\% test WER on Sl down to around 30\%).

\paragraph{Out-of-domain pre-training} We examine the out-of-domain pre-training setting using the Common Voice (CV) ASR corpus~\citep{ardila-etal-2020-common}. In contrast with the political domain oral speech in \vp, they are more fluent read speech of no copyright sentences (for example, Wikipedia articles). We adopt the few-shot phoneme recognition setup on CV v3 from~\citet{riviere2020unsupervised}, with which domain adaptation is limited during fine-tuning due to the small data volume --- it has 1-hour train set, 20-minute development set and 1-hour test set for 10 languages including 5 \vp~ones. We present the performance of VP-Mono-5K, VP-10K and VP-100K with the m-CPC~\citep{riviere2020unsupervised} and XLSR~\citep{conneau2020unsupervised} baselines in Table~\ref{tab:cv_eval_per}, where phone error rate (PER) is reported. The XLSR baselines share the same wav2vec 2.0 architecture as our models but are trained with in-domain CV data. VP-Mono-5K outperforms XLSR-Mono and XLSR-10 on all 5 \vp~languages (except for a tie on Es with XLSR-Mono). VP-100K outperforms XLSR-10 on 8 (9) out of the 10 languages. VP-100K (Large) overall performs competitively to XLSR-53, which leverages 52K-hour out-of-domain data in addition to the in-domain CV data. Notably, it outperforms XLSR-53 on Zh, which is covered by XLSR-53 but remote from the EU languages in VP-100K. This suggests the high generality of the speech representations VP-100K learned.

We also evaluate our multilingual model (VP-50K) under the normal setup (CV v5.1) and report test WER in Table~\ref{tab:wer_scores_main_cv}. 
They are compared with supervised baselines from  DeepSpeech-Polyglot\footnote{https://gitlab.com/Jaco-Assistant/deepspeech-polyglot}, which leverage extended CV train sets and several other corpora for training as well as LM for decoding. Our model outperforms the baseline with fine-tuning on the standard CV train set (a subset of the baseline's one), even when not using LM in decoding.

\paragraph{Out-of-language pre-training}
In the few-shot phoneme recognition setup (Table~\ref{tab:cv_eval_per}), VP-100K does not cover 5 of the 10 CV languages (Ky, Ru, Tr, Tt and Zh) in pre-training, but leverages data from 18 additional EU languages. It outperforms the in-domain in-language XLSR baselines on most of the uncovered languages (except Ky which is a remote central Asian language). Moreover, it performs more stably across all the 10 languages with a smaller variance (standard deviation) on PER.

\subsubsection{Self-Training for ASR and ST}
\label{sec:self_training}
Self-training~\citep{scudder1965probability} is a classical semi-supervised learning approach, where unlabeled data is equipped with pseudo-labels from a supervised model and then combined with labeled data for model training. We use the combination of EuroParl-ST~\citep{iranzo2020europarl} and CoVoST 2~\citep{wang2020covost} for both ASR and ST labeled data in 3 languages (directions). The former is created from 2009-2012 EP plenary sessions and hence has the same domain as \vp. The latter is based on Common Voice v4, which has different domain than \vp~and dominates the combined train set. We train Transformer \emph{Base}~\citep{NIPS2017_3f5ee243} supervised baselines and use 0.8K/3K-hour monolingual \vp~unlabeled data (from 2013-2020 sessions only to avoid overlaps with EuroParl-ST) to self-train Transformer \emph{Large} models. We upsample labeled data in self-training so that it has the same duration as the unlabeled one. We observe from Table~\ref{tab:st_self_training_eval} that self-training on \vp~improves both in-domain (``EP") and out-of-domain (``CV") performance with similar magnitude most of the time. For ST, self-training helps to narrow the gap between end-to-end models and the cascaded ones (more labeled data available) without the addition of expensive labeled data.

\subsection{Weakly Supervised ST}
We evaluate the quality of the weakly labeled ST data from our speech-to-speech alignment on the same benchmark as the self-training experiments. This also provides an indirect evaluation for our alignment pipeline since imprecise alignments hurt the ST label quality. 
We examine the performance of weakly supervised training as well as joint training using both labeled and weakly labeled data. We see from Table~\ref{tab:st_self_training_eval} that the former is on par with (or better than) the supervised baseline in the \vp~domain (``EP") with 0.3x-1.8x more training data than the baseline. Joint training brings substantial gains to both in-domain (``EP") and out-of-domain (``CV") performance, and it outperforms self-training. This suggests that our weakly labeled data (0.4K hours) is much more informative and efficient than the pseudo-labeled data (3K hours) when combined with labeled data.

\section{Related Work}

\paragraph{Multilingual speech corpora}
LibriLight~\citep{kahn2020libri} currently represents the largest scale unlabeled speech corpus but it is limited to English.
MLS~\cite{Pratap2020} is a recently released large-scale multilingual corpus of read speech in 8 languages, derived from LibriVox.
MAILABS\footnote{https://www.caito.de/2019/01/the-m-ailabs-speech-dataset} is also derived from Librivox and has about 1000 hours available in 9 languages.
While MLS and MAILABS are derived from audiobooks, VoxForge\footnote{http://www.voxforge.org} and Common Voice~\citep{ardila-etal-2020-common} gather data via crowd-sourcing. VoxForge collected data in about 15 different languages with about 300 hours of speech in total; Common Voice currently supports 60 languages for a total of 7327 validated hours available.
The CMU Wilderness dataset~\citep{black2019cmu} collects readings from the New Testament, with 700 different languages available.
IARPA Babel program\footnote{https://www.iarpa.gov/index.php/research-programs/babel} collected data for 24 languages, mostly
from conversational telephone speech. The dataset is however not released and under an open license, and focused on low-resource languages, with labeled data ranging between 25 to 65 hours per language.

\paragraph{Speech-to-Text and Speech-to-Speech Translation}

Apart from machine translation~\citep{koehn2005europarl}, the European Parliament open data has fostered the development of corpora for speech-to-text translation and for simultaneous interpretation. EuroParl-ST~\citep{iranzo2020europarl} is a multilingual speech-to-text translation corpus with translations between 6 European languages (En, Fr, De, Es, It and Pt). Similarly, EPIC~\citep{bendazzoli2005approach} is derived from the European Parliament with simultaneous interpretation speeches in Italian, English and Spanish. CIAIR~\citep{tohyama2004ciair} and STC~\citep{shimizu2014collection} are simultaneous interpretation corpora between English and Japanese with a total of about 180 hours for the former, while the latter is currently unavailable for download. The MaSS dataset~\citep{zanon-boito-etal-2020-mass} also provides speech to speech alignments for about 8k utterances across 8 languages, for a total of about 23h of speech.


\section{Conclusion}

In this paper, we introduce a large-scale multilingual speech corpus, \vp, for representation learning, semi-supervised learning and interpretation. \vp~provides the largest open unlabeled speech data to date, which has broad applications including unsupervised pre-training and self-training. \vp~is also the first corpus for large amounts of open speech-to-speech interpretation data.
We provide \vp~ASR baselines and validate the versatility of \vp~unlabeled data in semi-supervised learning under challenging out-of-domain settings. The corpus is available
at \url{https://github.com/facebookresearch/voxpopuli}.

\section{Acknowledgements}
We thank Gabriel Synnaeve, Tatiana Likhomanenko, Jade Copet, Vineel Pratap, Jiatao Gu and Alexis Conneau for helpful discussions on the project.

\section{Ethical Considerations}
We acknowledge the European Union (EU) for creating and publishing the materials used by \vp.
We will add citations as well as acknowledgements in our release.
We paid the market price to transcription vendors for the human annotations we collected. \vp~includes all available speeches from the 2009-2020 EP events without any selections on the topics or speakers. The speech contents represent the standpoints of the speakers in the EP events, many of which are EU officials.

\bibliography{acl2021}
\bibliographystyle{acl_natbib}

\end{document}